\documentclass{article}
\usepackage{spconf,amsmath,graphicx}

\usepackage{amsmath,graphicx}
\usepackage{lipsum} 
\usepackage{multirow}
\usepackage{amsmath}
\DeclareMathOperator*{\argmax}{argmax}
\usepackage{amssymb}

\title{ELASTIC CRFS FOR OPEN-ONTOLOGY SLOT FILLING}
\name{Yinpei Dai$^\dagger*$\thanks{$^*$These authors contributed equally to this work. Supported by NSFC 61473168, Ministry of Education and China Mobile
joint funding MCM20170301.}, Yichi Zhang$^\dagger*$, Hong Liu$^\dagger$, Zhijian Ou$^\dagger$, Yi Huang$^\ddagger$, Junlan Feng$^\ddagger$}
\address{$^\dagger$Speech Processing and Machine Intelligence (SPMI) Lab, Tsinghua University, China\\
$^\ddagger$ China Mobile Research Institute\\
\small {dyp16, zhangyic17}@mails.tsinghua.edu.cn, ozj@tsinghua.edu.cn, {fengjunlan}@chinamobile.com
}

\begin{document}
%
\maketitle
\begin{abstract}
Slot filling is a crucial component in task-oriented dialog systems  that is used to parse (user) utterances into semantic concepts called slots.
	An ontology is defined by the collection of slots and the values that each slot can take.
	The most widely used practice of treating slot filling as a sequence labeling task suffers from two main drawbacks.
	First, the ontology is usually pre-defined and fixed and therefore is not able to detect new labels for unseen slots.
	Second, the one-hot encoding of slot labels ignores the correlations between slots with similar semantics, which makes it difficult to share knowledge learned across different domains.
	To address these problems, we propose a new model called elastic conditional random field (eCRF), where each slot is represented by the embedding of its natural language description  and modeled by a CRF layer. New slot values can be detected by eCRF whenever a language description is available for the slot. In our experiment, we show that eCRFs outperform existing models in both in-domain and cross-domain tasks, especially in predicting unseen slots and values.
\end{abstract}
\begin{keywords}
Open ontology, slot filling, conditional random
fields, dialog systems
\end{keywords}
\section{Introduction}
\label{sec:intro}

\begin{figure}[tp]
	\includegraphics[width=\linewidth]{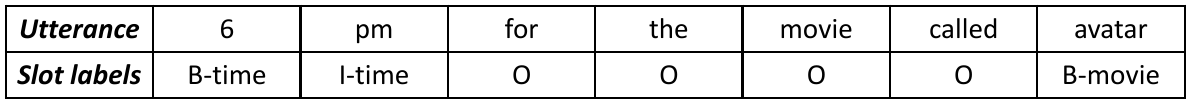}

	\caption{{An example of slot filling in the movie domain.}}
	\label{fig:slot-filling-example}
\vspace{-0.5cm}
\end{figure}

Slot filling \cite{Wang2005Spoken, Gr2015Using} is a crucial component in task-oriented dialog systems and parses (user) utterances into semantic concepts in terms of a set of named entities called slots.
The example in Figure \ref{fig:slot-filling-example} contains the slots \texttt{time} and \texttt{movie}.
In parsing, some span in the utterance is identified as the slot value for some slot; e.g., here, \textit{``6 pm''} is marked as the slot \texttt{time}.
An ontology, which describes the scope of semantics that the dialog system can process, is defined by the collection of slots and the values that each slot can take.
A widely  used practice for slot filling is to introduce   IOB tags \cite{Ramshaw1999Text} and assign a label to each token in the utterance. 
A label, e.g., \texttt{{B-time}}, is a combination of the slot name and one of the IOB tags.
These labels are then used to identify the values for different slots from the utterance.
In this manner, slot filling is treated as a sequence labeling task, as illustrated in Figure \ref{fig:slot-filling-example}, for which the two dominant classes of methods are based on recurrent neural networks (RNNs) \cite{Wang2005Spoken} and conditional random fields (CRFs) \cite{Lafferty2001Conditional}, respectively.
This practice has been widely employed for slot filling \cite{Gr2015Using,Liu+2016} and many other similar sequence labeling problems \cite{Sang2002Introduction}. However, this practice suffers from two drawbacks.

First, currently, most slot-filling methods are unable to predict new labels for unseen slots. The ontology is usually pre-defined and fixed. It is difficult to accommodate new semantic concepts (slots) in slot filling.
However, users may often add new semantic concepts in a domain and dialog systems are expected to work across an increasingly wide range of domains.
Thus, it is highly desirable for slot-filling models to be able to handle new slots, whether in-domain or cross-domain, with the least expense being incurred after training on a certain domain.
In this paper, we are interested in developing such open-ontology slot filling, which means that the collection of slots and values is open-ended for slot filling.
Second, in current slot-filling models \cite{Liu+2016, liu2015recurrent}, slot labels are generally encoded as one-hot vectors.
However, slot labels are not merely discrete classes. There are natural language descriptions for each slot, e.g., the description ``number of people'' for the slot \texttt{\#people}.
This one-hot encoding ignores the semantic meanings and relations for slots, which are implicit in their natural language descriptions and useful for slot filling.

There are prior efforts to address the above two drawbacks.
The difficulty of transferring between domains could be partly alleviated with multi-task learning \cite{Rastogi2017Scalable, Hakkani2016Multi, Mrk2015Multi}, by performing joint learning on multiple domains. 
Practically, varying only the last output layer for different domains and sharing the parameters of the rest layers has   shown to be a successful approach \cite{Aaron2016Domain}.
In this approach, the slot-filling model can leverage all available multi-domain data and transfer them to handle those slots with sparse training data.
However, basically, this multi-task learning approach is unable to predict labels for zero-shot slots (namely those slots that are unseen in training data and whose values are unknown).
It can be seen that this difficulty is also related to the drawback of   one-hot encoding   slot labels, which hinders the exploitation of semantic relations and shared statistical properties between different slots.
A recent work \cite{Bapna2017Towards} proposes   utilizing slot label descriptions towards zero-shot slot filling  by introducing slot encodings from natural language descriptions.
Basically, they use RNN-based sequence labeling, taking the slot encoding vector as an additional conditional input and outputting the IOB tags in each position. Sequence labeling is carried out independently for all slots.
Though yielding promising results, there are two shortcomings. First, independent sequence labeling may make conflicting predictions. Second, interactions between slots are ignored in sequence labeling.

CRFs have been shown to be one of the most successful approaches for sequence labeling, especially for capturing the interactions between labels.
A widely used method is to implement a CRF layer on top of features generated by a RNN \cite{Wang2005Spoken}. These recent neural CRFs are different from conventional CRFs, which mainly use discrete indicator features.
However, these recent CRFs still work with a closed set of labels. In this paper, we propose a novel neural CRF model, called elastic CRF (eCRF), for open-set sequence labeling, by leveraging label descriptions inspired from \cite{Bapna2017Towards}. The key idea of eCRFs is to use slot descriptions to create semantically meaningful IOB tags \cite{Ramshaw1999Text}, which are further used for a new calculation of potential functions in the CRF framework. Compared to traditional fixed IOB tags in original CRFs, our eCRFs are able to process new slots unseen during training without retraining the model. Such flexibility is the motivation for calling it an ``elastic'' CRF model.

The eCRFs are powerful models for open-ontology slot filling.
Intuitively, the node potentials of eCRFs combine the neural features of both the utterance and the slot descriptions, and the edge potentials model the interactions between different slots.
In the experiments, we make use of the Google simulated dataset \cite{shah2018building}, and re-split the dataset according to the in-domain task and the cross-domain task, which focus on the challenge of handling unseen values and unseen slots, respectively.
The results show that eCRFs significantly outperform  not only a BiLSTM baseline but also the concept tagger (CT) in \cite{Bapna2017Towards} for both tasks, especially in predictions of unseen slots and values.

In Section \ref{sec:related_works}, we discuss related work. The new eCRF model is detailed in Section \ref{sec:eCRF}. Section \ref{sec:dataset} describes the dataset and task formulations. Section \ref{sec:experiments} presents the experiments, followed by the conclusion in Section \ref{sec:conclusion}.

\section{Related Work}

\label{sec:related_works}
One line of related work is zero-shot slot-filling learning \cite{Larochelle2014Zero, rojas2018deep}. The term \textit{{open ontology}}
 referred in this paper is  a different name for   zero-shot slot filling in spoken language understanding (SLU) for dialog systems. 
Zero-shot learning has been applied in various of SLU tasks. The authors of \cite{Chen2016Zero} leverage  the intent embeddings to detect new intent labels which are not included in the training data. Additionally, \cite{Bapna2017Towards} exploits the slot label descriptions to parse the novel semantic frames for domain scaling and \cite{Zhao2018Zero} extends the natural language generation module to generalize the responses into an unseen domain via latent action matching. The authors  of \cite{shah-etal-2019-robust} propose    utilizing both the slot description and a small number of examples of slot values to enhance model robustness. In \cite{lin-etal-2021-leveraging}, the authors focus  on multi-turn zero-shot slot filling in  conversation.
These studies have utilized the natural language descriptions of the labels, and by constructing the semantic encoder to take the label descriptions as inputs, any new labels in the testing phrase can still be predicted by the model. 
Our eCRFs also use this semantic encoder structure. However, unlike processing each label description separately in \cite{Bapna2017Towards}, eCRFs are trained and tested by jointly exploiting all possible slot descriptions at one time.
Thus, they could capture relations between slot labels and relieve the burden of adjusting the oversampling ratio.

Another line of related work is models for slot filling. CRFs have been extensively applied in traditional slot-filling tasks \cite{chen2017improving,Xu2014Convolutional, dai2022cgodial, si2023spokenwoz,dai2021preview,zhang2020improved,dai2018tracking}, but are restricted by a fixed set of labels. 
With the progress of deep learning, state-of-art slot-filling methods usually utilize BiLSTM networks \cite{Hakkani2016Multi,Kurata2016Leveraging,Vu2016Bi, dai2020survey}. Extended models, such as encoder--decoder \cite{Liu+2016} and  memory network \cite{Chen2016End} designs, are explored. More recently, \cite{liu-etal-2020-coach} proposes a coarse-to-fine approach (Coach) for cross-domain slot filling, which detects the value span boundary first and then predicts the specific fine types for the slot entities. With the advance of pre-trained models \cite{devlin2018bert}, there are also many work \cite{li2019unified, gao2020machine, ijcai2021-0550, tseng2021transferable, dai2020learning} that adapt the well-studied machine reading comprehension (MRC) framework to solve open-ontology slot filling or using pre-tained dialogue models to generate slot labels \cite{he2022galaxy, he2022space, he2022unified,wang2022task, mi2021towards}.
Motivated by the BiLSTM-CRF architecture \cite{Ma2016End,chen2017improving,Lample2016Neural}, our eCRFs combine  the representation power of deep neural networks and dependency modeling ability of CRFs, together with a newly designed potential function.


\section{Proposed Model}
\label{sec:eCRF}

\begin{figure}[t]
	%
		\includegraphics[width=8.8cm]{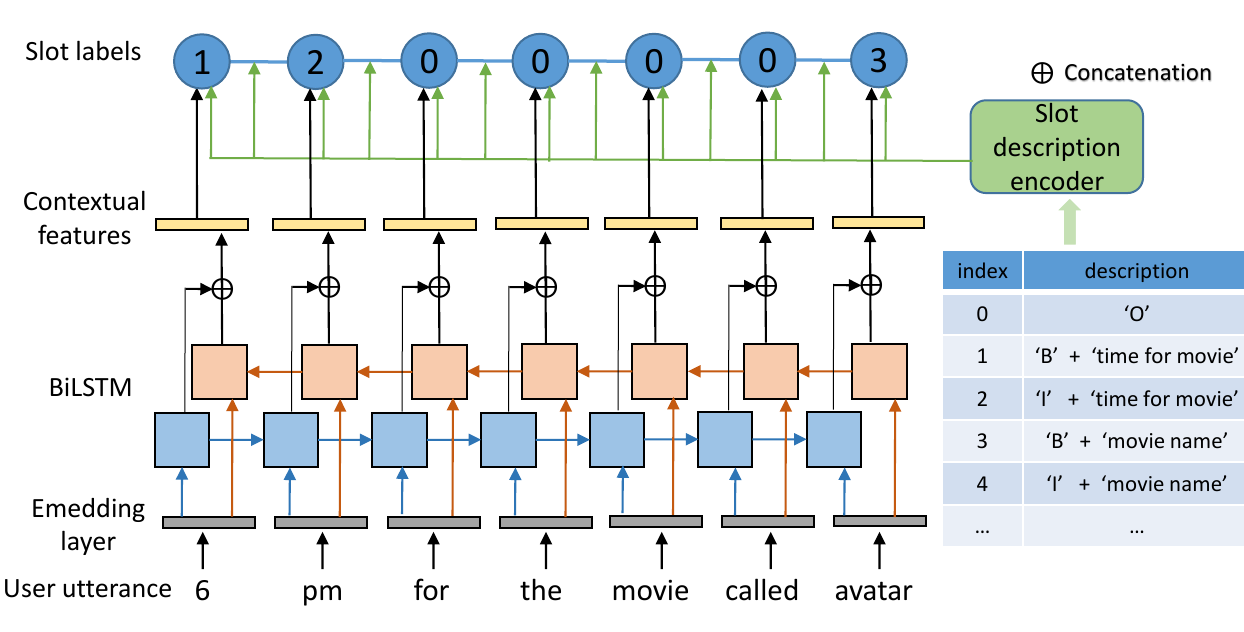}
	 
		\caption{The architecture of the elastic CRF (eCRF) model.}
		\label{fig:eCRF_model}
  \vspace{-0.5cm}
	\end{figure}

Our new model presents an extension from existing neural CRFs \cite{Ma2016End, Lample2016Neural}. Existing neural CRFs in many other sequence labeling tasks are restricted by a fixed set of labels, e.g., \textit{PERSON, LOCATION, ORGANIZATION, MISC} in the name entity recognition (NER) task, and thus can not be applied for open-ontology slot filling. 
To overcome this shortcoming, we propose a novel framework called elastic conditional random field (eCRF), which consists of three parts. (1) A {slot description encoder} is employed to encode the slot descriptions into semantic embeddings, then (2) a BiLSTM is used to extract contextual neural features, and finally (3) the outputs of both the slot description encoder and the BiLSTM are combined to define a novel potential function in the CRF. The main framework of eCRF is illustrated in {Figure}  \ref{fig:eCRF_model} and each part is detailed in the following subsections.


\subsection{Slot Description Encoder}

Let $X = (x_1,x_2,...,x_n)$ denote the input user utterance  and $D^{i}=(d^{i}_1,d^{i}_2,...)$ denote the description  
of slot $s^{i}$. {In our experiment, slot descriptions are simple complementary phrases, e.g., `number of people' for the slot \texttt{\#people}, `theatre name' for the slot \texttt{theatre\_name}, but other richer expression can be used.} 
The goal of our task is to find all possible text spans in $X$ as values for each $s^{i}$. 
We adapted the IOB tagging scheme as in \cite{Ramshaw1999Text}.
Traditionally, the IOB tags are made up  three type, `$B$', `$I$', and `$O$', which indicate  the beginning position of a value span, the intermediate and ending positions of the value span and the rest position belonging to no values. To be specific, if a word is predicted to have the `$B$' tag or multiple words are predicted to have `$B, I,..,I$' tags, the word span is the value of a slot. 
Instead of using a combination of the slot name and one of the IOB tags as in  Figure \ref{fig:slot-filling-example}, we used the combination of the slot description and one of the IOB tags in order to leverage the semantic meanings of slots. As shown in Figure \ref{fig:eCRF_model},
the slot description encoder takes all slot descriptions as input, and outputs are distributed representations for all possible combinations of the IOB tags and the slot descriptions, such as `$O$',  `$B+D^{1}$', `$I+D^{1}$', `$B+D^{2}$', `$I+D^{2}$', $...$. The set of these new combined slot labels is denoted as $\mathcal{S}$. We use indexes of these labels to suggest the corresponding positions within the utterance. For example, in Figure \ref{fig:eCRF_model},   `6', `pm' and `avatar' are predicted as the positions of `$B+$ time for movie',  `$I+$ time for movie' and `$B+$ movie name', which means that `6 pm' is the value of slot \texttt{movie\_time} and  `avatar' is the value of slot \texttt{movie\_name}.
A function $e(\cdot) \in \mathbb{R}^{d}$ is used to denote the output vector from the slot description encoder as follows:
\begin{align}
	\vspace{-0.6cm}
	e(B+D^{i})&=FC\left(f(D^{i}) \oplus emb(B)\right)\\
	e(I+D^{i})&=FC\left(f(D^{i}) \oplus emb(I)\right)\\
	e(O)&=FC\left(\overrightarrow{\textbf{0}} \oplus emb(O)\right)
\end{align}
where $FC(\cdot)$ denotes a  one-hidden-layer fully connected network and $f(\cdot)$ denotes an encoder that maps the descriptions into semantic embeddings. In this paper, we use a simple averaging function
of all word embeddings in $D^{i}$ as in \cite{Bapna2017Towards}. $emb(\cdot)$ is an embedding lookup function for the IOB tags and $\oplus$ denotes the concatenation operation. Note that for $e(O)$, we use a zero vector $\overrightarrow{\textbf{0}}$ with the same size as the output vector of $f(\cdot)$ since the `$O$' tag should be independent of any $D^{i}$. 
A difference between our slot description encoder and that in \cite{Bapna2017Towards} is that we leverage the embeddings of the IOB tags  so that the dependencies between tags in different slot labels are modeled.

\begin{figure}[t] 
	 		\includegraphics[width=7.8cm]{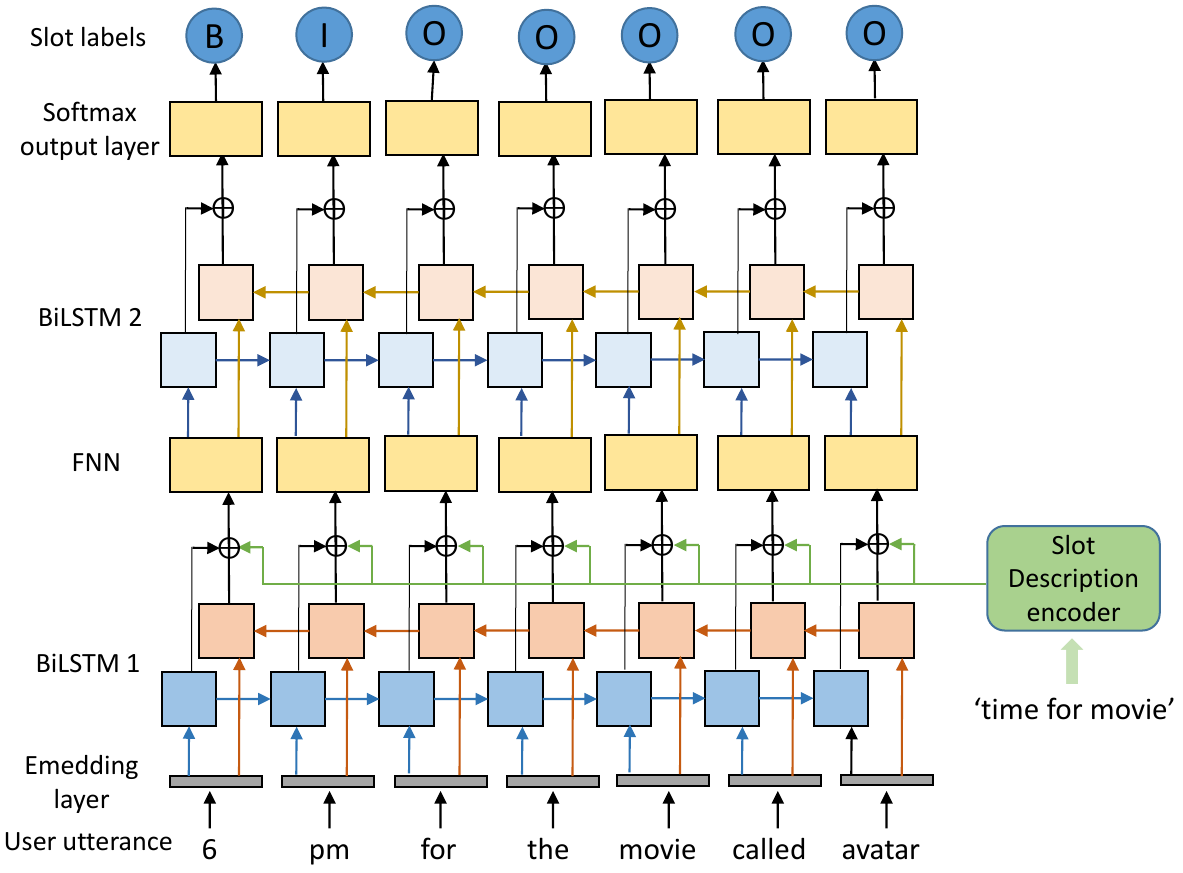}
	 
		\caption{{The architecture of the concept Tagging (CT) model.} \cite{Bapna2017Towards}} 
	 	\label{fig:google_model}
	    \vspace{-0.5cm}
\end{figure}

\subsection{BiLSTM Feature Extractor}
Bidirectional long short-term memory (BiLSTM) has been widely utilized in sequence models to capture the contextual semantic feature of input sentences \cite{Ma2016End,chen2017improving}. In eCRF, we also exploit BiLSTMs to extract the contextual neural features. Through concatenating the hidden states from both forward and backward passes, we acquire the distributed representations of contextual features $H=(h_1,h_2,...h_n) $, in which each $h_i \in \mathbb{R}^{d}$.

\vspace{-0.2cm}
\subsection{Elastic CRF (eCRF) Labeler}
\label{eCRF-labeler}
Let $Y=(y_1,y_2,...y_n)$ denote the output sequence of slot labels, where $y_i\in \mathcal{S}$. Then the potential function of our elastic neural CRF is defined as follows:
\vspace{-0.2cm}
\begin{align}
	\Psi(Y, W) = \sum_{i=1}^n{e(y_i)^Th_i} + 
	\sum_{i=1}^{n-1}{e(y_i)^TWe(y_{i+1})}
\end{align}
where $W\in \mathbb{R}^{d\times d}$ is a learnable matrix. The potential function consists of two items. The first term, called the node potential, calculates semantic similarity of the slot descriptions and the extracted contextual features.
The second term, called the edge potential, captures interactions between the slot labels through a bilinear calculation. Then, the likelihood of eCRF is defined as follows:
\vspace{-0.2cm}
\begin{align}
	p(Y|X,D)=\frac{\exp(\Psi(Y,X))}{\sum_{Y'}\exp(\Psi(Y',X))}
\end{align}
\vspace{-0.3cm}

The eCRF is trained by conditional maximum likelihood (CML), and we used Viterbi decoding for inferences as follows:
\vspace{-0.1cm}
\begin{align}
	\hat{Y}=\argmax_{y_1',...,y_n'\in\mathcal{S}} p(y_1',...,y_n'|X,D) 
\end{align}

In our experiment, we employed the pre-train trick \cite{David2015Structured} to speed up model learning. Namely, we first masked the edge potential term and trained only with the node potential term for a certain number of training steps, and then added the edge potentials in training. More details can be found in Section \ref{sec:setup}.

\section{Dataset and Tasks}

\label{sec:dataset}
In the experiments, we used the recent Google simulated dataset {(}accessed from  https://github.com/google-research-datasets/simulated-dialogue on 1 June 2018{)} as our main dataset. It is collected by the machines talking to machines (M2M) self-play schema \cite{shah2018building}. Two domains, restaurant and movie, were chosen. There are two common slots, i.e., \texttt{time} and \texttt{date}, in both domains, and an around 40\% out-of-vocabulary (OOV) rate in the test sets. However, since this dataset was not originally built for the open-ontology slot filling, the number of unseen values in the testing set is very limited. In order to properly use this dataset for the study, we designed two different tasks, the in-domain task and the cross-domain task, and accordingly re-split the whole dataset into new training and testing sets.

In the in-domain task, we aimed to evaluate various models for handling unknown values given all known slots. 
For each domain, we re-split the whole dataset by fixing the ratio between the number of types of values in training and testing.
Suppose the sets of all values occurred in the training set and testing set are $V_{train}$ and $V_{test}$, respectively; we defined the value ratio between training and testing as $|V_{train}|: |V_{test}-V_{train}|$. Three value ratios were chosen for model evaluations, that is, 75:25, 50:50 and 25:75. 

For the cross-domain task, we aimed to evaluate various models for handling unknown slots. Similar to the zero-shot multi-domain learning \cite{Bapna2017Towards}, we trained the model on one domain and evaluated it on the other domain. The common slots of the two domains are treated as known slots while the other slots were treated as unknown slots. 


After determining the training and testing sets, a validation set is randomly extracted from the training set, satisfying two conditions: (1) the ratio between the total number of utterances in the new training set and validation set is 4:1, and (2) around 50\% of the validation set contains unseen slots or values with respect to the new  training set. 
In this way, a reasonable validation set is constructed so that model training can be monitored for stopping for open-ontology prediction.

\begin{table*}[h]
    \centering

\scalebox{0.7}{
    \begin{tabular}{|c|c|c|c|c|c|c|c|c|c|c|}
        \hline
        \multirow{2}{*}{{Domain}} &
        \text{ {Value-Ratio}} &
        \multicolumn{3}{c|}{{Average Accuracy for Known Values}} &
        \multicolumn{3}{c|}{{Average Accuracy for Unknown Values}} &
        \multicolumn{3}{c|}{{Average Accuracy for Total Values}} \\
        \cline{3-11}
        & { {Train: Test}} & {BT} & {CT} & {eCRF} &  {BT} & {CT} & {eCRF} & {BT} & {CT} & {eCRF} \\
        \hline
        \multirow{3}{*}{sim-R} & 75:25 & 
        0.959$\pm$0.020 & {\textbf{0.993$\pm$0.005}} & 0.982$\pm$0.007 &
        0.555$\pm$0.122 & 0.753$\pm$0.108 & \textbf{0.791$\pm$0.047} &
        
        0.765$\pm$0.069 & 0.862$\pm$0.060 & \textbf{0.875$\pm$0.026} \\
        \cline{2-11}
        & 50:50 & 
        0.968$\pm$0.017 & \textbf{0.994$\pm$0.002} & 0.984$\pm$0.011 &
        
        0.361$\pm$0.083 & 0.474$\pm$0.066 & \textbf{0.618$\pm$0.058} &
        
        0.639$\pm$0.048 & 0.677$\pm$0.042 & \textbf{0.754$\pm$0.035} \\
        \cline{2-11}
        & 25:75 & 
        0.967$\pm$0.041 & \textbf{0.999$\pm$0.001} & 0.985$\pm$0.009 & 
        
        0.365$\pm$0.034 & 0.441$\pm$0.035 & \textbf{0.516$\pm$0.036} & 
        
        0.554$\pm$0.016 & 0.575$\pm$0.030 & \textbf{0.624$\pm$0.027} \\
        
            \hline
        \multirow{3}{*}{sim-M} & 75:25 & 
        0.951$\pm$0.034 & 0.982$\pm$0.005 & \textbf{0.984$\pm$0.003} & 
        
        0.843$\pm$0.009 & 0.876$\pm$0.066 & \textbf{0.905$\pm$0.011} & 
        
        0.914$\pm$0.018 & 0.930$\pm$0.037 & \textbf{0.953$\pm$0.005} \\
        
        \cline{2-11}
        & 50:50 & 
        0.941$\pm$0.028 & \textbf{0.982$\pm$0.009} & 0.975$\pm$0.017 & 
        
        0.655$\pm$0.024 & 0.723$\pm$0.076 & \textbf{0.841$\pm$0.024} & 
        
        0.803$\pm$0.014 & 0.840$\pm$0.040 & \textbf{0.910$\pm$0.017} \\
        \cline{2-11}
        & 25:75 & 
        0.948$\pm$0.024 & \textbf{0.991$\pm$0.003} & 0.988$\pm$0.005 & 
        
        0.519$\pm$0.034 & 0.611$\pm$0.030 & \textbf{0.682$\pm$0.035} & 
        
        0.662$\pm$0.027 & 0.718$\pm$0.021 & \textbf{0.784$\pm$0.023} \\
        
        \hline
\end{tabular}}
 \caption{Results for the in-domain tasks: average exact matching accuracies for known values, unknown values and total values for three models. Models are BiLISM tagging (BT) model, concept tagging (CT) model \cite{Bapna2017Towards} and elastic CRF (eCRF). Sim-R and sim-M are the domains of restaurant and movie respectively. 
		For each domain, three ratios between the number of types of values in training and testing are chosen to re-split the whole dataset to train models. Bold numbers mean the best results among three compared models.}
	\label{tab:in-domain}
\end{table*}

\begin{table*}[h]

	\label{tab:cross-domain}
	\scalebox{0.75}{
		\begin{tabular}{|c|c|c|c|c|c|c|c|c|c|c|}
			\hline
			{{Train}} &
			{{Test}} &
			\multicolumn{3}{c|}{{Average Accuracy for Known Slots} } &
			\multicolumn{3}{c|}{{Average Accuracy for Unknown Slots}} &
			\multicolumn{3}{c|}{{Average Accuracy for Total Slots}} \\
			\cline{3-11}
			{Domain} & {Domain} & {BT} & {CT} & {eCRF} &  {BT} & {CT} & {eCRF} & {BT} &{ CT} & {eCRF} \\
			\hline
			sim-M & sim-R & 
			0.980$\pm$0.025 & 0.974$\pm$0.009 & \textbf{{0.988$\pm$0.004}} & 
			0.136$\pm$0.045 & 0.121$\pm$0.077 & \textbf{0.243$\pm$0.009} & 
			0.502$\pm$0.036 & 0.491$\pm$0.044 & \textbf{0.566$\pm$0.007} \\
		\hline
			sim-R & sim-M & 
			0.814$\pm$0.064 & 0.915$\pm$0.013 & \textbf{0.926$\pm$0.024} & 
			0.165$\pm$0.040 & 0.246$\pm$0.017 & \textbf{0.377$\pm$0.031} & 
			0.508$\pm$0.035 & 0.599$\pm$0.006 & \textbf{0.667$\pm$0.020} \\
			\hline
	\end{tabular}}
  \caption{Results for the cross-domain tasks: average exact matching accuracies for values from known slots, unknown slots and total slots on test domain for three models. Bold numbers mean the best results among three compared models.}
\end{table*}

\section{Experiments}
\label{sec:experiments}
\subsection{Baselines}

In this paper, we compare our eCRF model with the concept tagging model proposed in \cite{Bapna2017Towards}  and a simple BiLSTM-based tagging model.

As shown in Figure 	\ref{fig:google_model}, the Concept tagging (CT) model employs a slot description encoder that takes the slot descriptions as input without the IOB tags. A one-layer BiLSTM is used to extract the contextual features of user utterances. 
The contextual features and the description encoder outputs are  concatenated and sent to a feedforward neural network (FNN).
This is followed by another one-layer BiLSTM. 
Finally, a softmax layer is used to calculate the distribution over slot labels. 
Since the slot descriptions are already used as conditional inputs, the output slot label set only consists of three labels, i.e., `$I$', `$B$', `$O$'. In both training and testing, the descriptions of each slot are iteratively fed into the model and evaluated separately.

The BiLSTM tagging (BT) model is a simplified version of the CT model, created  by removing the second BiLSTM layer. As shown in the following experimental results, this second BiLSTM layer plays an important role in transforming the contextual features and slot label features, which largely improves the performance.
\vspace{-0.2cm}

\subsection{Experimental Setup}
\label{sec:setup}
In our experiment, the vocabulary size is 1264. We use the open tool {(} accessed from https://github.com/stanfordnlp/GloVe on 25 October 2015{)} to train the GloVe embeddings on the whole dataset. The dimension of all word embeddings and the IOB tags are set as 50. The concatenated hidden size of all BiLSTMs are set as 100. The FNNs in the CT and BT models consist of one hidden layer with 100 units. 
For the pre-training of eCRFs, the edge potential is added in training after 2000 steps. All models are trained with the Adam \cite{Kingma2014} optimization method with a learning rate of 0.001.  Early-stopping is employed on the validation set to prevent over-fitting. For both the CT and BT models, we leveraged oversampling, which   sets the ratio of positive and negative samples as 1:1 and trains the model with a minibatch size of 10. 
For eCRFs, we set the minibatch size as 1. All the codes were implemented with Tensorflow\cite{Abadi2016TensorFlow}.  

\subsection{In-Domain Task Results}
As described in Section \ref{sec:dataset}, for the in-domain tasks, we re-organized the whole dataset into three different new datasets with increasing prediction difficulties, by setting the value ratios between training and testing as 75:25, 50:50 and 25:75.
Table \ref{tab:in-domain} shows the average exact-matching accuracies for known values, unknown values, and total values on the testing set for each model.



The results demonstrate that eCRFs clearly outperform the BT models in all conditions. 
Though slightly worse than the CT models on known values, eCRFs achieve much better results than the CT models in terms of accuracies for unknown values.
And the superiority becomes larger as the value-ratio in testing set becomes higher.
Therefore, in terms of accuracies for total values, eCRFs achieve the best overall performances.

\subsection{Cross-Domain Task Results}
For the cross-domain tasks, we train models on one domain and test on the other. The common slots such as  \texttt{time}, \texttt{date} are treated as known slots while the rest as unknown slots, such as \texttt{theatre\_name}, \texttt{restaurant\_name}. The evaluation metrics are the average exact-matching accuracies for values from known slots, unknown slots and total slots on the target domain. As shown in Table \ref{tab:cross-domain}, eCRFs outperform other models in all conditions. In the cross-domain tasks, although there are some overlapping between the known slots on the two domains, the user utterances are different in expressing those slots and values. These results demonstrate that our eCRFs have greater generalization ability.

\begin{figure}[h]
	\begin{minipage}[b]{1.0\linewidth}
		 
		\includegraphics[width=0.92\linewidth]{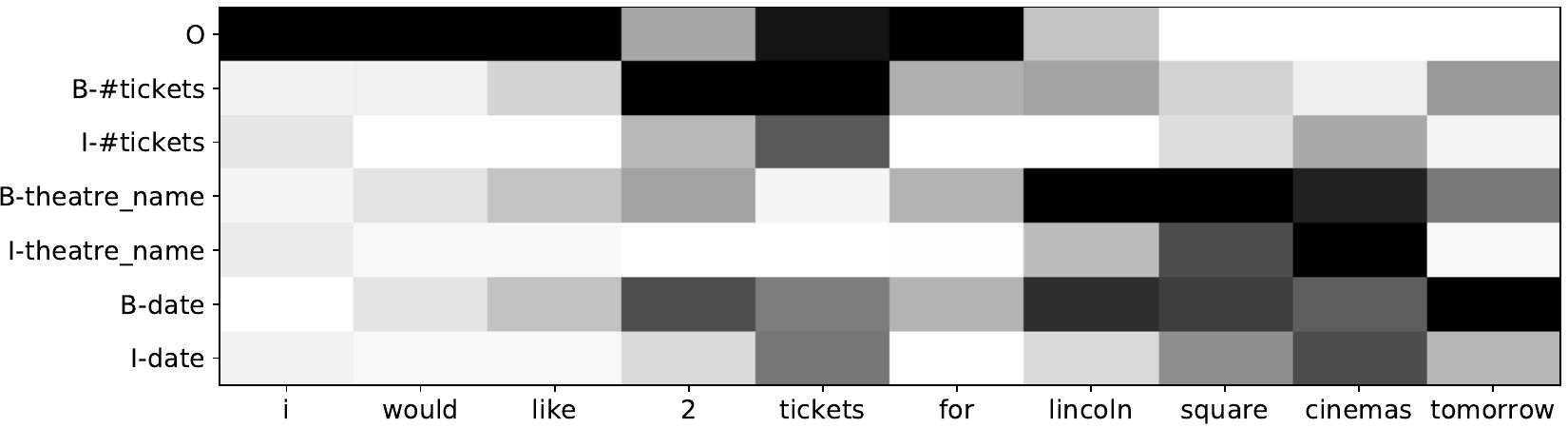}
		\vspace{-0.5cm}
		\caption{{Potential scores with only node potentials in eCRFs for the cross-domain task. The darker the color, the higher the potential score.}}\medskip 
		\label{fig:eCRF-node-potentials}
	\end{minipage}
	\vspace{-1.0cm}
\end{figure}

\begin{figure}[h]
	\begin{minipage}[b]{1.0\linewidth}
	 
		\includegraphics[width=0.85\linewidth]{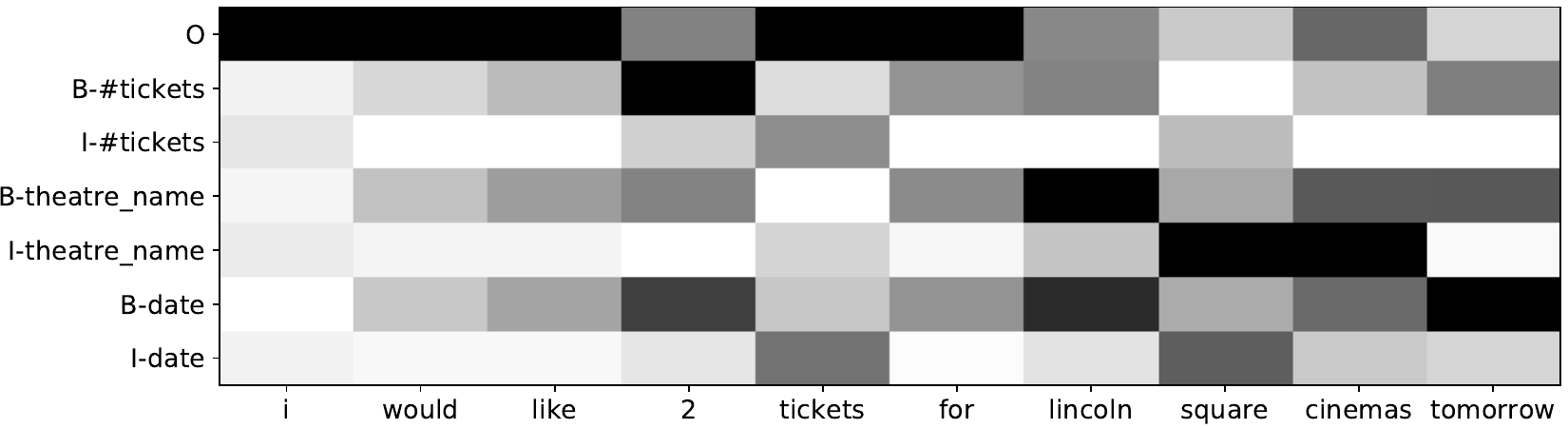}
		\vspace{-0.5cm}
		\caption{Potential scores with both node and edge potentials in eCRFs for the cross-domain task.}\medskip
		\label{fig:eCRF-all-potentials}
	\end{minipage}
	\vspace{-1.0cm}
\end{figure}

\begin{figure}[h]
	\begin{minipage}[b]{1.0\linewidth}
		 
		\includegraphics[width=0.85\linewidth]{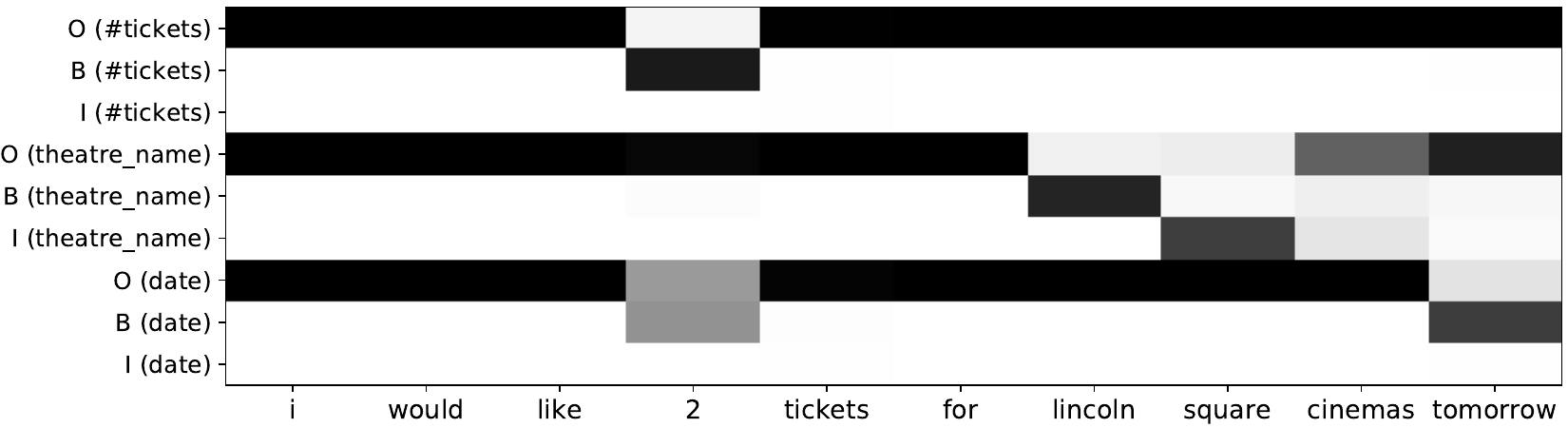}
		\vspace{-0.5cm}
		\caption{Probabilities of the IOB labels for each slot in the CT model.}\medskip
		\label{fig:CT-probability}
	\end{minipage}
	\vspace{-1.0cm}
\end{figure}

Figures \ref{fig:eCRF-node-potentials}--\ref{fig:CT-probability} show the prediction results for the same utterance on the movie domain with the eCRF and CT models.
Figure \ref{fig:eCRF-node-potentials} illustrates the predicted scores with only node potentials for eCRFs, while Figure \ref{fig:eCRF-all-potentials} gives the predicted scores with both node and edge potentials. 
It can be seen that the boundaries of slot labels for some slots are mistakenly placed in Figure \ref{fig:eCRF-node-potentials}, e.g., the value ``\textit{lincoln square cinemas}" for the unknown slot \texttt{theatre\_name} is falsely predicted as two values ``\textit{lincoln}" and ``\textit{square cinemas}".  
When taking both node and edge potentials into account, correct predictions are obtained for all the three slots, as shown in Figure \ref{fig:eCRF-all-potentials}. 
The output probabilities of slot labels for the CT model are shown in Figure \ref{fig:CT-probability}. Although the CT model gives the right prediction for the known slot \texttt{date} and unknown slot \texttt{\#tickets}, it mistakenly predicts the value for the unknown slot \texttt{theatre\_name} as ``\textit{lincoln square}", as it fails to learn the semantic relations between slot labels.


\section{Conclusions}
\label{sec:conclusion}
In this paper, we propose a novel model, the elastic conditional random field (eCRF), for open-ontology slot-filling task. The natural language descriptions of slots and (user) utterances are encoded into the same semantic embedding space to implement the node and edge potentials. We recompose the Google simulated dataset and demonstrate that eCRFs achieve better performances in both in-domain tasks and cross-domain tasks than existing models. 

There are interesting future works to further enhance the parsing ability and adaptation capacity of eCRFs: (1) encoding the descriptions of more semantic labels including the intent labels, domain labels and action labels for better generalization and (2) upgrading the CRF architecture with a slot label language model that can capture long-range dependencies between labels.


\bibliographystyle{IEEEbib}
\bibliography{strings,refs}

\end{document}